\documentclass[10pt,twocolumn,letterpaper]{article}

\usepackage{cvpr}
\usepackage{times}
\usepackage{epsfig}
\usepackage{graphicx}
\usepackage{amsmath}
\usepackage{amssymb}
\usepackage{enumitem}
\usepackage{algorithm}
\usepackage[noend]{algpseudocode}
\usepackage{subfigure}

\newcommand{\model}{GDAN}

\usepackage[pagebackref=true,breaklinks=true,letterpaper=true,colorlinks,bookmarks=false]{hyperref}

\cvprfinalcopy 

\def\cvprPaperID{1062} 

\ifcvprfinal\fi
\begin{document}

\title{Generative Dual Adversarial Network for Generalized Zero-shot Learning}


\author{
He Huang\textsuperscript{1} \hspace{0.5cm} Changhu Wang\textsuperscript{2} \hspace{0.5cm} Philip S. Yu\textsuperscript{1} \hspace{0.5cm} Chang-Dong Wang\textsuperscript{3}\\
\textsuperscript{1}University of Illinois at Chicago, USA\\
\textsuperscript{2}ByteDance AI Lab, China\\
\textsuperscript{3}Sun Yat-sen University, China\\
{\tt\small \{hehuang, psyu\}@uic.edu \hspace{0.1cm} wangchanghu@bytedance.com  \hspace{0.1cm} changdongwang@hotmail.com}
}

\maketitle

\begin{abstract}
   This paper studies the problem of generalized zero-shot learning which requires the model to train on image-label pairs from some seen classes and test on the task of classifying new images from both seen and unseen classes. Most previous models try to learn a fixed one-directional mapping between visual and semantic space, while some recently proposed generative methods try to generate image features for unseen classes so that the zero-shot learning problem becomes a traditional fully-supervised classification problem. In this paper, we propose a novel model that provides a unified framework for three different approaches: $visual\rightarrow semantic$ mapping, $semantic\rightarrow visual$ mapping, and metric learning. Specifically, our proposed model consists of a feature generator that can generate various visual features given class embedding features as input, a regressor that maps each visual feature back to its corresponding class embedding, and a discriminator that learns to evaluate the closeness of an image feature and a class embedding. All three components are trained under the combination of cyclic consistency loss and dual adversarial loss. Experimental results show that our model not only preserves higher accuracy in classifying images from seen classes, but also performs better than existing state-of-the-art models in classifying images from unseen classes. 
\end{abstract}


\section{Introduction}
Deep learning models have achieved great success in image classification tasks~\cite{ILSVRC15ImageNet}, and these models are so effective that they are comparable with humans. However, humans are much better at recognizing novel objects that they only saw a few times before or heard of but never saw. This is because deep learning models for image classification rely heavily on fully supervised training, and thus they require a lot of labeled data. As there are too many classes in the real world, it is almost impossible to collect enough labeled data for each class. In this case, these models are challenged by the task of recognizing images from classes that are unseen during training, which is also called the \emph{zero-shot learning} (ZSL) problem~\cite{akata2013ALE, lampert2009DAP-IAP}. In conventional zero-shot learning, the goal is to train an image classifier on a set of images from seen classes, and then test the trained model using images from unseen classes, where the sets of seen and unseen classes are totally disjoint, and that during testing the label space only contains unseen classes. However, this conventional setting is based on a strong assumption that in testing stage the model knows whether an image is from the set of seen or unseen classes, which is not realistic and thus not applicable in the real world. When the model receives a new image, it does not know whether it comes from seen or unseen classes, and thus it needs the ability to classify images from the combination of seen and unseen classes, which is called \emph{generalized zero-shot learning} (GZSL)~\cite{xian2018gbu, chao2016empirical}.  The main difference between zero-shot learning and generalized zero-shot learning is the label space during testing, which means that models designed for conventional zero-shot learning can still be applied in the generalized zero-shot learning setting. 

\begin{figure}[t]
\begin{center}
   \includegraphics[width=0.9\linewidth]{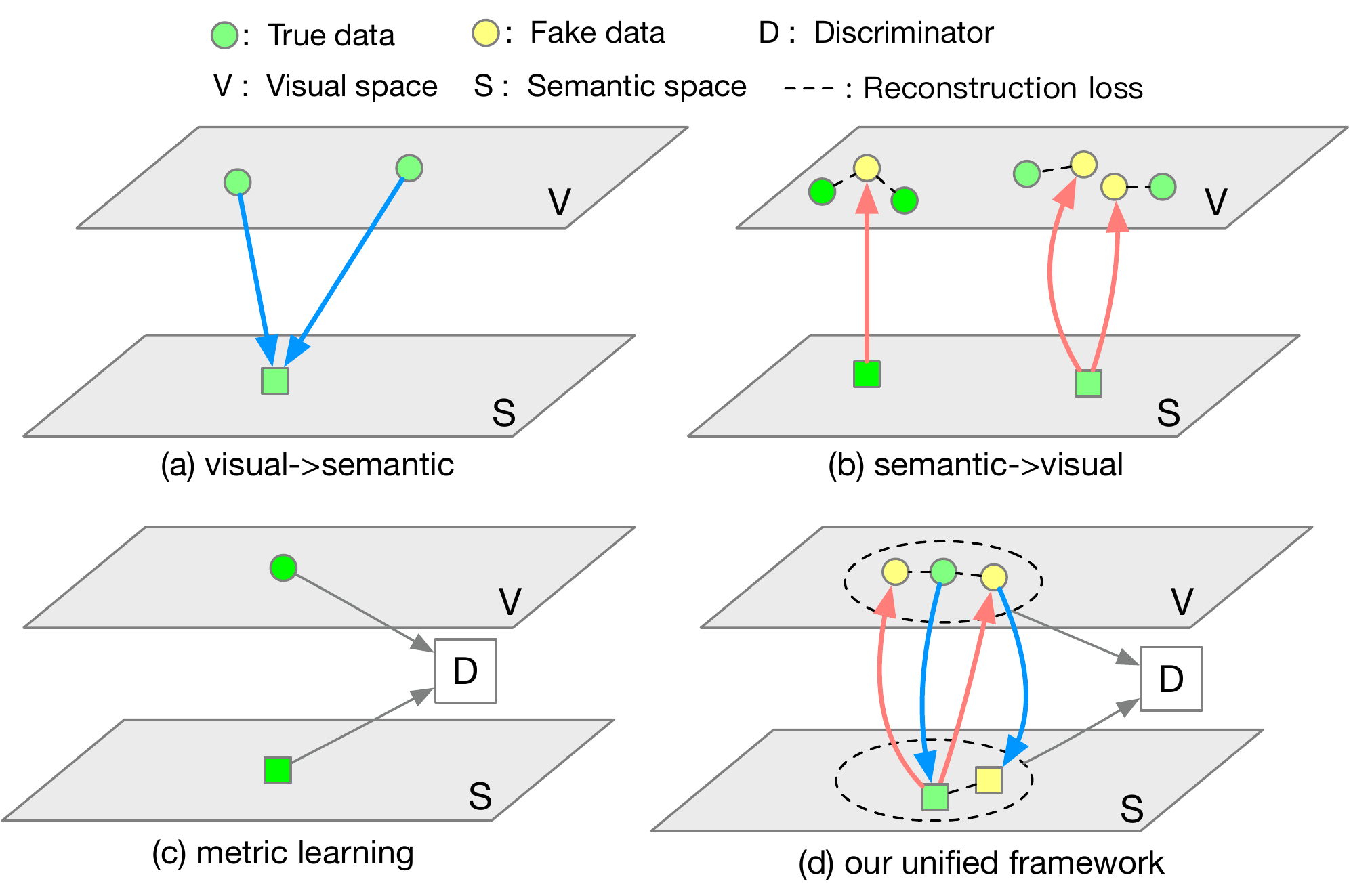}
\end{center}
   \caption{Four categories of zero-shot learning methods. (a) Embedding models that map visual features of the same class to the same feature in semantic space. (b) Models that map each class's semantic feature to features in the visual space. (c) Metric learning models that directly learn to evaluate the closeness of a pair of visual and semantic features. (d) Our proposed model that unifies the three methods using generative adversarial network and dual learning.}
\label{fig:category}
\end{figure}

A common strategy for tackling zero-shot learning is to map images and classes into the same latent space and then perform nearest neighbor search. Most existing methods project visual features to the semantic space spanned by class attributes, such as \cite{frome2013devise, xian2016LATEM, akata2013ALE, romera2015eszsl, akata2015SJE, kodirov2017SAE}, as is illustrated in Figure~\ref{fig:category}(a). However, as pointed out by \cite{shigeto2015ridge, dinu2014improving,zhang2017DEM}, using semantic space as shared latent space will suffer from the hubness problem, which means projecting high-dimensional visual features to low-dimensional space will further reduce the variance of features and the results may become more clustered as a hub. In order to mitigate this problem, \cite{shigeto2015ridge, dinu2014improving,zhang2017DEM} propose to project semantic features into the visual space, as illustrated in the left part of Figure~\ref{fig:category}(b). However, using a deterministic approach to map a class's semantic embedding to the visual space is still problematic, since one class label has numerous corresponding visual features. In contrast, some recent works \cite{mishra2017cvae-zsl, xian2018fganzsl, verma2018segzsl} propose to use generative methods that can generate various visual features conditioned on a semantic feature vector, as illustrated in the right part of Figure~\ref{fig:category}(b). Despite their effectiveness, their performance is limited by lacking either the ability to learn a bi-directional mapping between visual and semantic space, or an adversarial loss that acts as a more flexible metric to evaluate the similarity of features. Instead of manually choosing a common latent space, RelationNet~\cite{yang2018relation} proposes to learn a deep metric network which takes a pair of visual and semantic features as input and output their similarity, as illustrated in Figure~\ref{fig:category}(c). However, RelationNet~\cite{yang2018relation} fails to learn latent features of images and classes, nor does it support semi-supervised learning, and thus it is outperformed by some recent works~\cite{mishra2017cvae-zsl, xian2018fganzsl, verma2018segzsl}.

As the three different approaches have their different merits and  limitations, in this paper, we study the generalized zero-shot learning problem and propose a novel model named \textbf{G}enerative \textbf{D}ual \textbf{A}dversarial \textbf{N}etwork (\model) that combines $visual\rightarrow semantic$ mapping,  $semantic\rightarrow visual$ mapping as well as metric learning methods in a unified framework, as illustrated in Figure~\ref{fig:category}(d). More specifically, our model contains a generator network which is able to generate image features conditioned on class embedding, a regressor network that takes image features and output their class embedding (\ie semantic features), and a discriminator network that takes as input an image feature and a semantic feature and output a score indicating how well they match with each other. The generator and regressor learn from each other through a cyclic consistency loss, while both of them also interact with the discriminator through a dual adversarial loss. 

Our main contributions are summarized as follows:
\begin{itemize}[leftmargin=*,noitemsep,topsep=0pt]
    \item We propose a novel \textbf{G}enerative \textbf{D}ual \textbf{A}dversarial \textbf{N}etwork (\model) that unifies $visual\rightarrow semantic$, $semantic\rightarrow visual$ methods as well as metric learning for generalized zero-shot learning.
    \item In contrast to previous works in zero-shot learning, we design a novel dual adversarial loss so that the Regressor and Discriminator can also learn from each other to improve the model's performance.
    \item We conduct extensive experiments that demonstrate the ability of our proposed \model~ model in effectively classifying images from unseen classes as well as preserving high accuracy for seen classes on four widely used benchmark datasets.
    \item We conduct component analysis to show that the combination of three components in our model actually helps each of them in achieving better performance than individual ones. Visualization of synthetic samples from unseen classes also demonstrate the effective generation power of our model. Our code will also be available online.
\end{itemize}

\section{Related Work}

\textbf{Generative Adversarial Network}\hspace{0.1cm}
Generative Adversarial Network (GAN)~\cite{goodfellow2014GAN} was originally proposed as a method for image synthesis \cite{goodfellow2014GAN, radford2015dcgan} and has achieved state-of-the-art results. However, GAN is also known for its instability in training and suffers the mode collapse problem~\cite{arjovsky2017wgan, gulrajani2017wgan-gp}. In order to mitigate these problems and improve the quality of synthetic samples, many methods have been proposed. 
WGAN~\cite{arjovsky2017wgan} and WGAN-GP~\cite{gulrajani2017wgan-gp} propose to optimize GAN on an approximate Wasserstein distance by enforcing 1-Lipschitz smoothness. LS-GAN~\cite{mao2017lsgan} offers a simple but effective solution by replacing the cross-entropy loss of GAN with a least square loss that push the scores of real and fake samples to different decision boundaries so that the gradient never vanishes even when two distributions are totally disjoint. Our model also includes a GAN, which is realized by combining a feature generator and a discrminator. One major difference between our model and traditional GAN is that our model contains a non-generative component, \ie a regressor network, that interacts with the discriminator through an extra adversarial loss.

\textbf{Dual Learning}\hspace{0.1cm}
Dual learning has been shown to be effective in neural machine translation (NMT)~\cite{he2016duallearning} by training a prime task and a dual task together, where the dual task is the inverse task of the prime task. For example, in neural machine translation, the prime task may be $English\rightarrow French$, and then the dual task is $French\rightarrow English$. Dual learning has also been applied to other computer vision tasks like semantic segmentation~\cite{luo2017deep} and image-to-image translation~\cite{zhu2017cyclegan, yi2017dualgan}. Our work is related to CycleGAN~\cite{zhu2017cyclegan} and DualGAN~\cite{yi2017dualgan} in that we borrow the cyclic consistency loss from them. However, those two models require two generative networks which make them unable to be directly applied in generalized zero-shot learning, since each class has a fixed semantic representation and a generative network is not suitable for $visual\rightarrow semantic$ mapping as it may generate very diverse semantic features. Thus we need a novel architecture to incorporate cyclic consistency in zero-shot learning.

\textbf{(Generalized) Zero-shot Learning}\hspace{0.1cm} 
There are a few recent works that tackles the (generalized) zero-shot learning problem, which are closely related to our work. CVAE-ZSL~\cite{mishra2017cvae-zsl} proposes to use conditional variational autoencoder (CVAE)~\cite{sohn2015cvae} to generate samples for unseen classes, but it uses the vanilla version of CVAE which suffers from the prior collapse problem~\cite{zhao2018unsupervised}. SE-GZSL~\cite{verma2018segzsl} is another recent work that utilize variational autoencoder in generalized zero-shot learning. Although SE-GZSL~\cite{verma2018segzsl} also has a regressor that is similar to our model, it lacks the dual adversarial loss we have. In contrast, the discriminator of our model can learn a flexible metric to evaluate the relation between an image feature and a class embedding, and the regressor of our model can also learn from the discriminator through the adversarial loss. f-CLSWGAN~\cite{xian2018fganzsl} applies GAN to generate image features conditioned on class attributes, but it does not have the ability to map image features back to class attributes. RelationNet~\cite{yang2018relation} tries to learn a deep metric to evaluate the compatibility between an image feature and an semantic feature, which is similar to the discriminator in our model, while our model also has the ability to generate samples for unseen classes and infer  semantic embedding from visual features.

\section{Proposed Model}
In this section, we first formally define the generalized zero-shot learning problem, give an overview of our proposed model, and then introduce each part of our model in details.

\subsection{Problem Definition and Notations}
In this paper, we study the generalized zero-shot learning problem. Specifically, let the training data (including validation) be defined as $\mathcal{S}=\{ (\mathbf{v}, y, \mathbf{s}_y )| \mathbf{v} \in \mathcal{V}^s, y \in \mathcal{Y}^s, \mathbf{s}_y \in \mathcal{A} \}$, where $\mathbf{v}$ is an image's feature produced by a pre-trained neural network, $\mathcal{V}^s$ is the set of image features from seen classes, $y$ is the label of image feature $\mathbf{v}$, $\mathcal{Y}^s$ is the set of labels for seen classes, $\mathbf{s}_y$ is the attribute vector (semantic embedding) for class $y$. Similarly, we can define the test set as $\mathcal{U}=\{ (\mathbf{v}, y, \mathbf{s}_y )| \mathbf{v} \in \mathcal{V}^u, y \in \mathcal{Y}^u, \mathbf{s}_y \in \mathcal{A} \}$, where  $\mathcal{V}^u$ represent the set of image features from unseen classes, $\mathcal{Y}^u$ represents the set of labels for unseen classes, and that $\mathcal{Y}^u \cap \mathcal{Y}^s = \O$. The goal of generalized zero-shot learning is to learn a classifier  $f : \mathbf{v} \rightarrow \mathcal{Y}^u \cup \mathcal{Y}^s$, where $\mathbf{v}$ is the visual feature of an image from either seen or unseen classes.

\subsection{Model Overview}

\begin{figure}[!ht]
\begin{center}
   \includegraphics[width=0.7\linewidth]{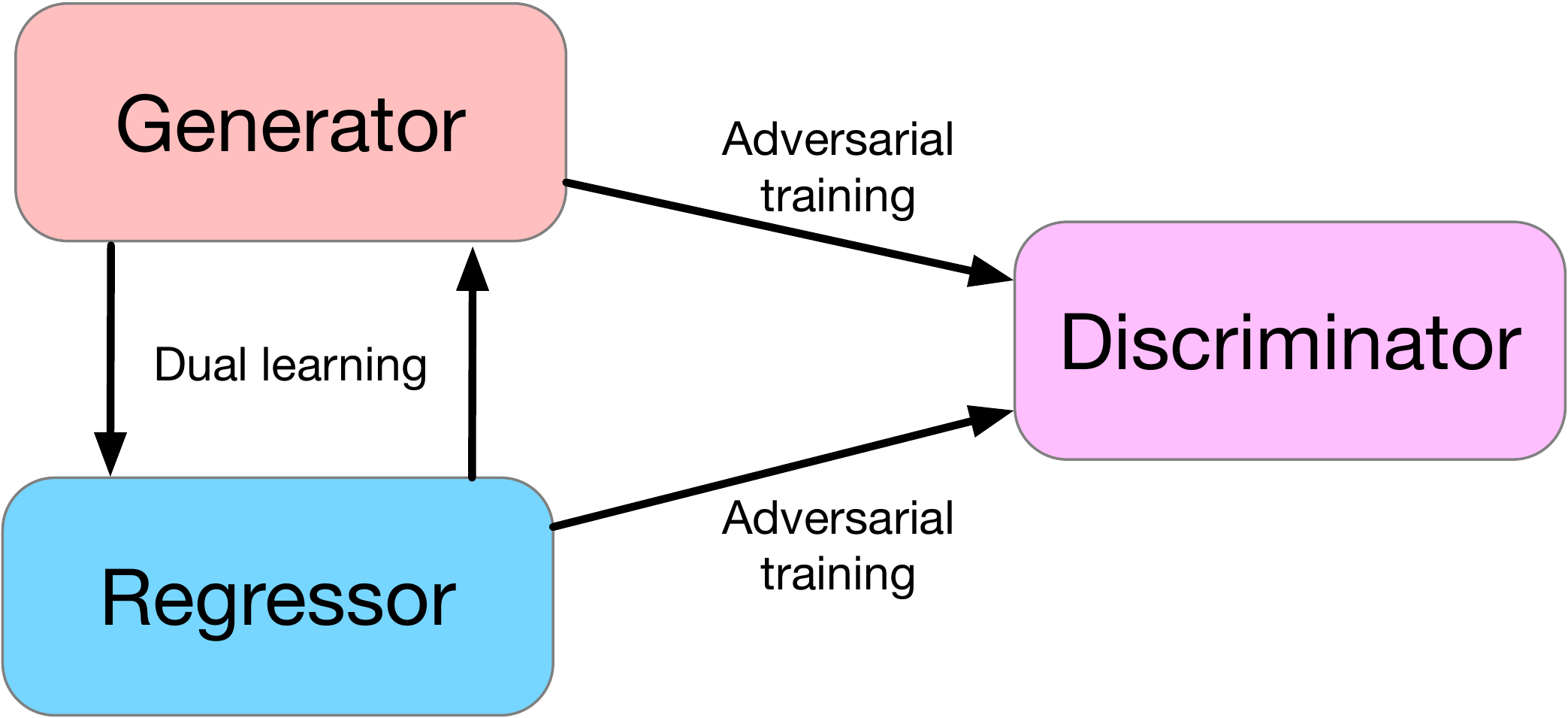}
\end{center}
   \caption{Overall framework of the proposed \model~model, where black arrows indicate data flow.}
\label{fig:framework}
\end{figure}

The overall framework of our proposed \textbf{G}enerative \textbf{D}ual \textbf{A}dversarial \textbf{N}etwork (\textbf{\model}) is illustrated in Figure \ref{fig:framework}. There are three components in our model, a Generator, a Regressor and a Discriminator. The core component of \model~ is the Generator network that can generate various visual features conditioned on certain class labels. Along with the Generator, we have a Regressor network that acts as a deep embedding function, which tries to map each visual feature back to its corresponding class's semantic feature. The Generator and the Regressor network together form a dual learning framework, so that they can learn from each other through a cyclic consistency loss. In addition, we have an extra Discriminator network that measures the similarity of a visual-textual feature pair, and it interacts with the other two networks through a dual adversarial loss. It should be noted that any of the three components is able to perform generalized zero-shot learning, where the Generator represents the $semantic\rightarrow visual$ methods (\eg \cite{mishra2017cvae-zsl,verma2018segzsl,xian2018fganzsl}), the Regressor represents the $visual\rightarrow semantic$ methods (\eg \cite{annadani2018psr, chen2018SPAEN}), and the Discrminator represents metric learning approach such as RelationNet~\cite{yang2018relation}. Our model provides a unified framework of all three different approaches so as to make use of their respective advantages and achieves better results for zero-shot image classification.  

\begin{figure*}[t]
\begin{center}
   \includegraphics[width=0.9\linewidth]{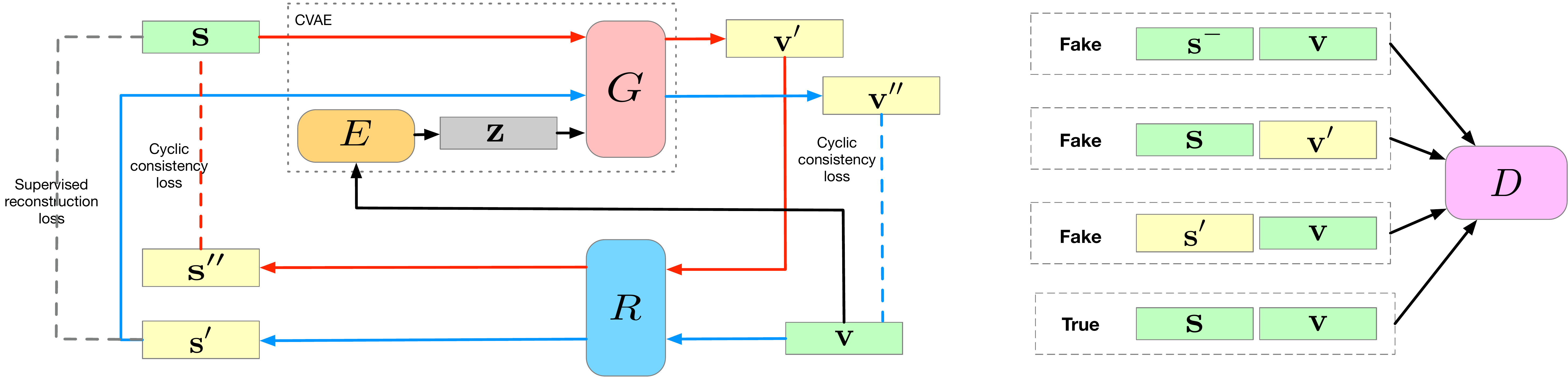}
\end{center}
   \caption{Detailed illustration of our proposed \model~model. $E$ and $G$ are the encoder and decoder/generator of the CVAE, $R$ represents the Regressor network, and $D$ represents the Discriminator for adversarial training. The CVAE and the Regressor interact with each other through a cyclic consistency loss, while they both learn from the Discriminator through a dual adversarial loss. The CVAE loss is not illustrated in the figure for clarity, and the Regressor has an additional supervised reconstruction loss.}
\label{fig:model}
\end{figure*}

\subsection{Feature Generation Network}
Our model's framework is very flexible, which means we can choose any generative model as the Generator. The simplest one can be a feed-forward neural network that takes as input the concatenation of a class embedding as well as a noise vector  randomly sampled from normal Gaussian distribution, which is widely used in Generative Adversarial Nets (GANs)~\cite{goodfellow2014GAN}. However, this naive model lacks the ability to infer noise vectors from image features. 
Thus, here we resort to Conditional Variational Autoencoder (CVAE)~\cite{sohn2015cvae}, which was proposed as a generative method that maps a random noise vector $\mathbf{z} \in R^{d_z}$ drawn from $P(\mathbf{z}|\mathbf{x, c})$ to a data point $\mathbf{x}$ in the data distribution conditioning on the context feature $\mathbf{c}$. 
A CVAE consists of two parts, an encoder $P_{E}(\mathbf{z}|\mathbf{x,c})$ that maps a data point $\mathbf{x}$, along with its class feature $c$, to its latent feature $\mathbf{z}$, and a decoder/generator $P_{G}(\mathbf{x}|\mathbf{z,c})$ that maps a latent vector to a data point.  The training objective of CVAE is to minimize the following loss function:
\begin{align}
    loss(\theta_E, \theta_G) = &~ \mathbb{E}_{P_{data}(\mathbf{x}, \mathbf{z}),P_{E}(\mathbf{z}|\mathbf{x,c})}[\log P_{G}(\mathbf{x}|\mathbf{z}, \mathbf{c})] \\ \notag 
    & - D_{KL}(P_{E}(\mathbf{z}|\mathbf{x,c})|| P(\mathbf{z})), \notag
\end{align}
where $D_{KL}(p||q)$ represents the Kullback-Leibler divergence between two distributions $p$ and $q$, and $P(\mathbf{z})$ is treated as a unit Gaussian distribution.

However, as pointed out in \cite{zhao2018unsupervised, zhao2017infovae, chen2016variational, makhzani2015aae}, CVAE is hard to train due to the posterior collapse problem. In order to mitigate this problem, we follow \cite{verma2018segzsl} to disentangle $\mathbf{z}$ from $\mathbf{c}$, so that the encoder only depends on $\mathbf{x}$ (\ie $P_G(\mathbf{z}|\mathbf{x})$), as illustrated in the top left of Figure~\ref{fig:model}. Also, as in \cite{makhzani2015aae}, we add an adversarial loss to help learning better CVAE, which we will discuss later. The loss of the conditional variational autoencoder (CVAE) used in our model is defined as follows:
\begin{align}
\label{eq:cvae}
    \mathcal{L}_{\textrm{CVAE}}(\theta_E, \theta_G) = &~ \mathbb{E}_{P_{data}(\mathbf{v}, \mathbf{s}),P_{E}(\mathbf{z}|\mathbf{v})}[\log P_{G}(\mathbf{v}|\mathbf{z,s})] \\ \notag 
    & - D_{KL}(P_{E}(\mathbf{z}|\mathbf{v})|| P(\mathbf{z})),
\end{align}
where $\mathbf{v}$ is an image's visual feature and $\mathbf{s}$ is the class embedding of $\mathbf{v}$.

Although generative models like VAE~\cite{kingma2013vae} and GAN~\cite{goodfellow2014GAN} have been proved to be effective in generating images, the output images of generative models are still blurry for images with complex objects, and the synthetic images often lack the sharpness to reveal the detailed attributes of objects, which are important for zero-shot learning~\cite{chen2018SPAEN}. Hence, instead of generating raw images, we train CVAE to generate visual feature $\mathbf{v}$, where the ground truth visual features are provided by image classification models pretrained on ImageNet~\cite{ILSVRC15ImageNet}.

\subsection{Regressor Network}
While the CVAE learns to generate visual features from semantic features, our model also has a Regressor network that performs the reverse task of mapping visual features back to their corresponding semantic features.
The CVAE and Regressor together form a dual learning framework so that they can learn from each other. In our case, the prime task is generating image features conditioned on class embedding, while the dual task is to transform image features back to their corresponding class embedding. 

As illustrated in the down left of Figure~\ref{fig:model}, the Regressor network $R$ takes two kinds of image features as input, where the first kind is the real image features $\mathbf{v}$ sampled from training data, and the second kind is the fake features $\mathbf{v}' = G(\mathbf{s,z})$ generated by the CVAE. With the paired training data $(\mathbf{v}, \mathbf{s})$, we can train the Regressor with a supervised loss:
\begin{align}
    \mathcal{L}_{\textrm{sup}}(\theta_R) = \mathbb{E}_{P_{data}(\mathbf{v}, \mathbf{s})}||\mathbf{s} - R(\mathbf{v})||^2_2.
\label{eq:sup}
\end{align}

In addition, the Regressor interacts with the CVAE with the following cyclic-consistency loss:
\begin{align}
    \mathcal{L}_{\textrm{cyc}}(\theta_G,\theta_E, \theta_R) &\ = \mathbb{E}_{P_{data}(\mathbf{v}, \mathbf{s}),P_E(\mathbf{z|v}) }[||\mathbf{v} - G(R(\mathbf{v}),\mathbf{z})||^2_2 \\ \notag 
    &\ + ||\mathbf{s} - R(G(\mathbf{s,z}))||^2_2],
\end{align}
where $G$ is the decoder/generator of CVAE and $P_E(\mathbf{z|v})$ is implemented by the encoder $E$ in Figure~\ref{fig:model}.

\subsection{Discriminator Network}
So far, the Generator and Regressor already combines $semantic\rightarrow visual$ and $visual\rightarrow semantic$ methods, but it still lacks the ability to learn from a flexible metric which can better evaluate the similarity of features. In order to incorporate metric learning, the third component of our model is a Discriminator network $D$ (as shown in the right of Figure \ref{fig:model}) that takes as input a visual-semantic feature pair $(\mathbf{v}, \mathbf{s})$ and output a compatibility score indicating the extent to which $\mathbf{v}$ belongs to the class characterized by $\mathbf{s}$. The Discriminator network learns how to evaluate the closeness of an image feature to a class, instead of using a predefined metric like L1/L2 distance or cosine similarity. 

In order to train the Discrimiantor to distinguish different types of fake data, inspired by Text-to-Image-GAN~\cite{reed2016txt2img}, we also train the Discriminator with two kinds of fake input, \ie $(G(\mathbf{s}, \mathbf{z}), \mathbf{s})$ and $(\mathbf{v}, \mathbf{s^-})$, where $s^-$ is a randomly sampled class's embedding and $s \neq s^-$. In addition, we add a novel third type of fake input generated by the Regressor, \ie  $(\mathbf{v},  R(\mathbf{v}))$. In this way, not only the CVAE but also the Regressor can learn from the Discriminator through adversarial training.

Since GANs are notoriously hard to train, there are many methods that try to stabilize the training process of GAN~\cite{goodfellow2014GAN,mao2017lsgan,arjovsky2017wgan,gulrajani2017wgan-gp}. In our model, we apply LS-GAN~\cite{mao2017lsgan} as the training method for adversarial loss for its simplicity and effectiveness.
Thus the adversarial loss for the Discriminator of our model can be defined as:
\begin{align}
\label{eq:dis}
    \mathcal{L}_{\textrm{adv}}(\theta_D) &\ = \mathbb{E}_{P_{data}(\mathbf{v,s})}[D(\mathbf{v,s}) - 1]^2  \\ \notag 
                &\ + \mathbb{E}_{P_{data}(\mathbf{v,s}),P_E(\mathbf{z}|\mathbf{v})}[D(G(\mathbf{s,z}),\mathbf{s})^2] \\ \notag 
                 &\ + \mathbb{E}_{P_{data}(\mathbf{v})}[D(\mathbf{v}, R(\mathbf{v}))^2]\\ \notag
                &\ + \mathbb{E}_{P_{data}(\mathbf{v,s}),P_{data}(\mathbf{s^-|s})}[D(\mathbf{v,s^-})^2],
\end{align}
where the first term is the loss for real samples $(\mathbf{v,s})$, the second term stands for the scores of features generated by the Generator, the third term stands for the scores of class embedding generated by the Regressor, and the last term is the control loss which trains the Discriminator to discriminate pairs of real image features and real negative class embedding. The negative sample $\mathbf{s}^-$ is randomly sampled from the set of training classes $\{\mathbf{s}_y^- | y \in \mathcal{Y}^s, \mathbf{s}_y^- \neq \mathbf{s} \}$. Here the discriminator $D$ tries to push the scores of real samples to $1$ and push the scores for generated samples to $0$, while the generator $G$ tries to push the scores of its synthetic samples to $1$.

Also, the adversarial loss for CVAE and Regressor can be defined as:
\begin{align}
     \mathcal{L}_{\textrm{adv}}(\theta_R) &\ = \mathbb{E}_{P_{data}(\mathbf{v})}[D(\mathbf{v}, R(\mathbf{v}))-1]^2 \\
    \mathcal{L}_{\textrm{adv}}(\theta_E,\theta_G) &\ = \mathbb{E}_{P_{data}(\mathbf{v,s}), P_E(\mathbf{z|v}))}[D(G(\mathbf{s,z}))-1]^2
\end{align}

\subsection{Full Objective and Training Procedure}

In adversarial training, the discriminator is training separately from the other two networks, while we train the CVAE and Regressor using an overall loss defined as:
\begin{align}
\label{eq:gen}
    \mathcal{L}(\theta_{G}, \theta_{E}, \theta_{D}, \theta_{R}) &\ = \mathcal{L}_{\textrm{CVAE}}(\theta_{G},\theta_{E}) +  \mathcal{L}_{\textrm{adv}}(\theta_{G},\theta_{E}) \notag  \\ 
    &\ + \lambda_1*\mathcal{L}_{\textrm{cyc}}(\theta_G, \theta_E,\theta_R) \notag \\
    &\ + \lambda_2*\mathcal{L}_{\textrm{sup}}(\theta_R) \notag \\
     &\ + \lambda_3*\mathcal{L}_{\textrm{adv}}(\theta_R),
\end{align}
where $\lambda_1$, $\lambda_2$ and $\lambda_3$ are hyper-parameters that assign weight on different parts of the overall loss.

The training procedure of our \model~ model is as follows: we first pretrain CVAE using Equation~\ref{eq:cvae}, and then train the whole model in an adversarial way using Equation~\ref{eq:dis} and Equation~\ref{eq:gen}. The implementation details will be provided in the experiment section.


\subsection{Evaluation Protocol}
Once the model has been trained, in order to predict the label for unseen classes, we can first generate new samples for each unseen class, and then combine those synthetic samples with other samples in the training data, after which we can train any new classifier based on this new dataset that contains samples for both seen and unseen classes. For fair comparison with other baselines, we just apply a simple 1-NN classifier for testing, which is used in most baselines.

\begin{table*}[t]
\begin{center}
\begin{tabular}{|c|c|c|c|c|c|c|}
\hline
Dataset & \#attributes & \begin{tabular}[c]{@{}c@{}}\#seen classes\\ (train+val)\end{tabular} & \begin{tabular}[c]{@{}c@{}}\#unseen\\ classes\end{tabular} & \begin{tabular}[c]{@{}c@{}}\#images\\ (total)\end{tabular} & \begin{tabular}[c]{@{}c@{}}\#images\\ (train+val)\end{tabular} & \begin{tabular}[c]{@{}c@{}}\#images\\ (test unseen/seen)\end{tabular} \\ \hline
\textbf{aPY}~\cite{farhadi2009apy} & 64 & 15+5 & 12 & 15339 & 5932 & 7924/1483 \\ \hline
\textbf{AwA2}~\cite{xian2018gbu} & 85 & 27+13 & 10 & 37332 & 23527 & 7913/5882 \\ \hline
\textbf{CUB}~\cite{wah2011cub_bird} & 312 & 100+50 & 50 & 11788 & 7057 & 2679/1764 \\ \hline
\textbf{SUN}~\cite{patterson2012sun} & 102 & 580+65 & 72 & 14340 & 10320 & 1440/2580 \\ \hline
\end{tabular}
\end{center}
\caption{Statistics of datasets.}
\label{tab:stats}
\end{table*}

\begin{table*}[]
\begin{center}
\begin{tabular}{|c|ccc|ccc|ccc|ccc|}
\hline
\textbf{Dataset}     & \multicolumn{3}{c|}{\textbf{SUN}} & \multicolumn{3}{c|}{\textbf{CUB}} & \multicolumn{3}{c|}{\textbf{AwA2}} & \multicolumn{3}{c|}{\textbf{aPY}} \\ \hline
\textbf{Methods}     & U      & S      & H      & U      & S      & H      & U       & S      & H      & U      & S      & H      \\ \hline
SSE~\cite{zhang2015SSE}        & 2.1    & 36.4   & 4.0    & 8.5    & 46.9   & 14.4   & 8.1     & 82.6   & 14.8   & 0.2    & 78.9   & 0.4    \\ 
LATEM~\cite{xian2016LATEM}       & 14.7   & 28.8   & 19.5   & 15.2   & 57.3   & 24.0   & 11.5    & 77.3   & 20.0   & 0.1    & 73.0   & 0.2    \\ 
ALE~\cite{akata2013ALE}         & 21.8   & 33.1   & 26.3   & 27.3   & 62.8   & 34.4   & 14.0    & 81.8   & 23.9   & 4.6    & 73.7   & 8.7    \\ 
DEVISE~\cite{frome2013devise}      & 16.9   & 27.4   & 20.9   & 23.8   & 53.0   & 32.8   & 17.1    & 74.7   & 27.8   & 4.9    & 76.9   & 9.2    \\ 
SJE~\cite{akata2015SJE}         & 14.7   & 30.5   & 19.8   & 23.5   & 52.9   & 33.6   & 8.0     & 73.9   & 14.4   & 3.7    & 55.7   & 6.9    \\ 
ESZSL~\cite{romera2015eszsl}       & 11.0   & 27.9   & 15.8   & 12.6   & 63.8   & 21.0   & 5.9     & 77.8   & 11.0   & 2.4    & 70.1   & 4.6    \\ 
SYNC~\cite{changpinyo2016SYNC}        & 7.9    & 43.3   & 13.4   & 11.5   & 70.9   & 19.8   & 10.0    & 90.5   & 18.0   & 7.4    & 66.3   & 13.3   \\ 
SAE~\cite{kodirov2017SAE}         & 8.8    & 18.0   & 11.8   & 7.8    & 54.0   & 13.6   & 1.1     & 82.2   & 2.2    & 0.4    & \textbf{80.9}   & 0.9    \\ 
DEM~\cite{zhang2017DEM}         & 34.3   & 20.5   & 25.6   & 19.6   & 57.9   & 29.2   & 30.5    & 86.4   & 45.1   & 11.1   & 79.4   & 19.4   \\ 
RelationNet~\cite{yang2018relation} & -      & -      & -      & 38.1   & 61.1   & 47.0   & 30      & \textbf{93.4}   & 45.3   & -      & -      & -      \\ 
PSR-ZSL~\cite{annadani2018psr}     & 20.8   & 37.2   & 26.7   & 24.6   & 54.3   & 33.9   & 20.7    & 73.8   & 32.3   & 13.5   & 51.4   & 21.4   \\ 
SP-AEN~\cite{chen2018SPAEN}      & 24.9   & 38.2   & 30.3   & 34.7   & \textbf{70.6}  & 46.6   & 23.3    & 90.9   & 31.1   & 13.7   & 63.4   & 22.6   \\ 
CVAE-ZSL~\cite{mishra2017cvae-zsl}    & -      & -      & 26.7   & -      & -      & 34.5   & -       & -      & \textbf{51.2}   & -      & -      & -      \\ \hline
\textbf{GDAN} & \textbf{38.1} & \textbf{89.9} & \textbf{53.4} & \textbf{39.3}   & 66.7   & \textbf{49.5}   & \textbf{32.1}  & 67.5   & 43.5   & \textbf{30.4} & {75.0} & \textbf{43.4} \\ \hline \hline
SE-GZSL*~\cite{verma2018segzsl}     & 40.9   & 30.5   & 34.9   & {41.5}   & 53.3   & 46.7   & {58.3}    & 68.1   & {62.8}   & -      & -      & -      \\ 
f-CLSWGAN*~\cite{xian2018fganzsl}   & 42.6   & 36.6   & 39.4   & 43.7   & 57.7   & 49.7   & -       & -      & -      & -      & -      & -      \\ \hline
\end{tabular}
\end{center}
\caption{Results of generalized zero-shot learning evaluated on four benchmark datasets. *Note that SE-GZSL~\cite{verma2018segzsl} trains an additional LinearSVC for testing, and that f-CLSWGAN~\cite{xian2018fganzsl} trains additional embedding models for testing, so their results may not be directly comparable with others.}
\label{tab:results}
\end{table*}

\section{Experiments}
In this section, we conduct extensive experiments on four public benchmark datasets under the generalized zero-shot learning setting. 

\subsection{Datasets and Settings}
We compare our \model~ model with several baselines on \textbf{SUN}~\cite{patterson2012sun}, \textbf{CUB}~\cite{wah2011cub_bird}, \textbf{aPY}~\cite{farhadi2009apy} and \textbf{AWA2}~\cite{xian2018gbu}. Among these datasets, \textbf{aPY}~\cite{farhadi2009apy} and \textbf{AWA2}~\cite{xian2018gbu} are coarse-grained and of small and medium size respectively, while  \textbf{SUN}~\cite{patterson2012sun} and \textbf{CUB}~\cite{wah2011cub_bird} are both medium fine-grained datasets. We follow the training/validation/testing split as well as the image and class features provided by \cite{xian2018gbu}. The statistics of these datasets is summarized in Table~\ref{tab:stats}. 

For image features and class embedding, we use publicly available features provided by \cite{xian2018gbu}. We also adopt the widely used \emph{average per-class top-1 accuracy} to evaluate the performance of each model, which is defined as follows:
\begin{align}
    Acc_\mathcal{Y} = \frac{1}{||\mathcal{Y}||}\sum_{c}^{||\mathcal{Y}||} \frac{\textrm{\# of correction predictions in }c}{\textrm{\# of  samples in } c}  
\end{align}

In generalized zero-shot learning setting, during test phrase we use images from both seen and unseen classes, and the label space is also the combination of seen  and unseen classes $\mathcal{Y}^s \cup \mathcal{Y}^u$. We want the accuracy of both seen and unseen classes to be as high as possible, thus we need a metric that can reflect the overall performance of a model. Since arithmetic mean can be significantly biased by extreme values, we follow \cite{xian2018gbu} and use harmonic mean instead. Let $Acc_{\mathcal{Y}^s}$ and $Acc_{\mathcal{Y}^u}$ denote the accuracy of images from seen and unseen classes respectively, the harmonic mean $H$ of seen and unseen accuracy is thus defined as:
\begin{align}
    H = \frac{2 * Acc_{\mathcal{Y}^u} * Acc_{\mathcal{Y}^s}}{Acc_{\mathcal{Y}^u} + Acc_{\mathcal{Y}^s}}
\end{align}

\subsection{Implementation Details}
 We implement the CVAE, Regressor and Discriminator of our model as feed-forward neural networks. The Encoder of CVAE has two hidden layers of 1200 and 600 units respectively, while the Generator of CVAE and the Discriminator is implemented with one hidden layer of 800 hidden units. The Regressor has only one hidden layer of 600 units. The dimension of noise vector $\mathbf{z}$ is set to 100 for all datasets. We use $\lambda_1=\lambda_2=\lambda_3=0.1$ and find that they generally work well. We choose Adam~\cite{kingma2014adam} as our optimizer, and the momentum is set to  $(0.9,0.999)$.The learning rate of Discriminator is set to $0.00001$, while the learning rate for CVAE and Regressor is $0.0001$. $d\_iter$ and $g\_iter$ are set to 1, which means all the modules of our model train with the same number of batches. We train on each dataset for 500 epochs, save the model checkpoints every 10 epochs, and then evaluate on the validation set to find the best one for testing. Our code is also available online\footnote{www.github.com/stevehuanghe/GDAN}.

\subsection{Results}

We compare our model with recent state-of-the-art methods on generalized zero-shot learning, and the results are shown in Table~\ref{tab:results}. 
Although f-CLSWGAN~\cite{xian2018fganzsl} also uses GAN for zero-shot learning, it trains additional embedding methods such as ~\cite{akata2013ALE,akata2015SJE,frome2013devise,xian2016LATEM}, while most baselines only use 1-NN for testing, thus its results may not be directly comparable with others. SE-GZSL~\cite{verma2018segzsl} is also closely related to our model, but it trains an additional LinearSVC for testing, so it is not directly comparable with other methods. For f-CLSWGAN~\cite{xian2018fganzsl} and SE-GZSL~\cite{verma2018segzsl}, we just copy and paste their results from the original papers~\cite{xian2018fganzsl, verma2018segzsl} in Table~\ref{tab:results} for reference.

As can be seen from Table~\ref{tab:results}, our model achieves a significant performance gain in SUN~\cite{patterson2012sun}. For both seen and unseen classes, our method achieves the highest accuracy among all baselines, and with a significant improvement in classifying images from seen classes, our method also outperform deep embedding models \cite{zhang2017DEM, annadani2018psr, chen2018SPAEN} and generative model such as \cite{mishra2017cvae-zsl} as well as metric learning model \cite{yang2018relation} by a large margin. This shows the benefits of putting $visual\rightarrow semantic$, $semantic\rightarrow visual$ and metric learning into one framework.

For CUB~\cite{wah2011cub_bird} dataset, our model achieves the highest accuracy for unseen classes, while our accuracy for seen classes is slightly lower than SP-AEN~\cite{chen2018SPAEN}. Still, we achieve the highest harmonic mean of 49.5\%, which is 2.5\% higher than the second place RelationNet~\cite{yang2018relation}. This again shows that our model maintains a good balance of predicting image from both seen and unseen classes, while previous method may not manage the trade-off as well as ours.

For AwA2~\cite{xian2018gbu}, our \model~ outperforms some recent methods like SP-AEN~\cite{chen2018SPAEN} and PSR-ZSL~\cite{annadani2018psr} in both unseen class accuracy and harmonic mean accuracy. Although DEM~\cite{zhang2017DEM} and RelationNet~\cite{yang2018relation} slightly outperforms our \model~ in harmonic mean accuracy by less than 1\%, \model~ achieves  a higher unseen class accuracy than them with a noticeable margin of 2.7\%.

According to \cite{chen2018SPAEN}, aPY~\cite{farhadi2009apy} has a much smaller cosine similarity (0.58) between the attribute variances of the disjoint train and test images than the other datasets (0.98 for SUN, 0.95 for CUB, 0.74 for AwA2), which means it is harder to synthesize and classify images of unseen classes. Although previous methods have relatively low accuracy for unseen classes, our performance gain is even higher with such a difficult dataset. Compared to all previous models, our \model~ achieves a higher accuracy for unseen classes by a large margin of 16\%, and still our model maintains a high accuracy for seen classes. From the results of previous models we can see that although they generally achieves very high accuracy for seen classes, they perform very poorly when predicting images from unseen classes, while our model achieves a good balance between seen and unseen classes, which give us the highest harmonic mean accuracy on aPY~\cite{farhadi2009apy}.

\subsection{Component Analysis}
In this section, we study the problem of whether the three components actually help each other in training, so we train them independently  to see how well they perform when they are alone. We train the CVAE component using Equation~\ref{eq:cvae} and evaluate using the same protocol as \model. For the Discriminator, we train it using Equation~\ref{eq:dis} but with  only two kinds of input, \ie $(\mathbf{v}, \mathbf{s})$ and $(\mathbf{v}, \mathbf{s}^-)$. During evaluation, for each image feature, we use the Discrimiantor to calculate its matching score with all class embedding and assign it to the class which has the largest score. As for the Regressor, we train it with Equation~\ref{eq:sup} and evaluate using 1-NN. We can also use the Discriminator and Regressor of our trained \model~ model and test their performance in generalized zero-shot learning, and we denote them as Discriminator-\model~ and Regressor-\model. We also train our \model~ without Discriminator (\model~w/o Disc) and without Regressor (\model~w/o Reg) respectively.

\begin{table}[!htb]
\begin{center}
\begin{tabular}{|c|c|c|c|c|}
\hline
\textbf{Datasets} & \textbf{SUN} & \textbf{CUB} & \textbf{AwA2} & \textbf{aPY} \\ \hline \hline
CVAE & 30.1 & 33.7 & 28.5 & 28.1 \\ 
Discriminator & 0.07 & 1.43 & 4.2 & 1.2 \\ 
Regressor & 0.1 & 3.4 & 3.8 & 10.1 \\ 
Discriminator-GDAN & 1.0 & 3.3 & 11.2 & 11.1 \\ 
Regressor-GDAN & 5.8 & 4.3 & 7.1 & 11.5 \\
GDAN w/o Disc & 37.3 & 38.2 & 31.4 & 29.6 \\
GDAN w/o Reg & 37.4 & 38.1 & 30.9 &  29.3\\
\textbf{GDAN} & \textbf{38.1} & \textbf{39.3} & \textbf{32.1} & \textbf{30.4} \\ \hline
\end{tabular}
\end{center}
\caption{Unseen class accuracy for component analysis.}
\label{tab:component}
\end{table}

The results of unseen class accuracy evaluated on four dataset are shown in Table~\ref{tab:component}. As we can see, the CVAE component alone is comparable to many baselines like PSR-ZSL~\cite{annadani2018psr} and SP-AEN~\cite{chen2018SPAEN}, while the Regressor and Discrimiantor are very weak on their own, which is reasonable since they are implemented as neural networks with only one hidden layer. If we train with only two of the components, as we can see from the second and third last row of Table~\ref{tab:component}, our model still have a major performance gain compared to other baselines, and the Regressor or Discriminator have similar effect on improving the performance of CVAE. 
In addition, if we train all three components together, we find out that not only the CVAE component generates better samples for unseen classes, but also the Discriminator and Regressor have a performance boost, which demonstrates the effectiveness of putting three components under a unified framework.

\begin{figure}[!ht]
\begin{center}
   \includegraphics[width=0.75\linewidth]{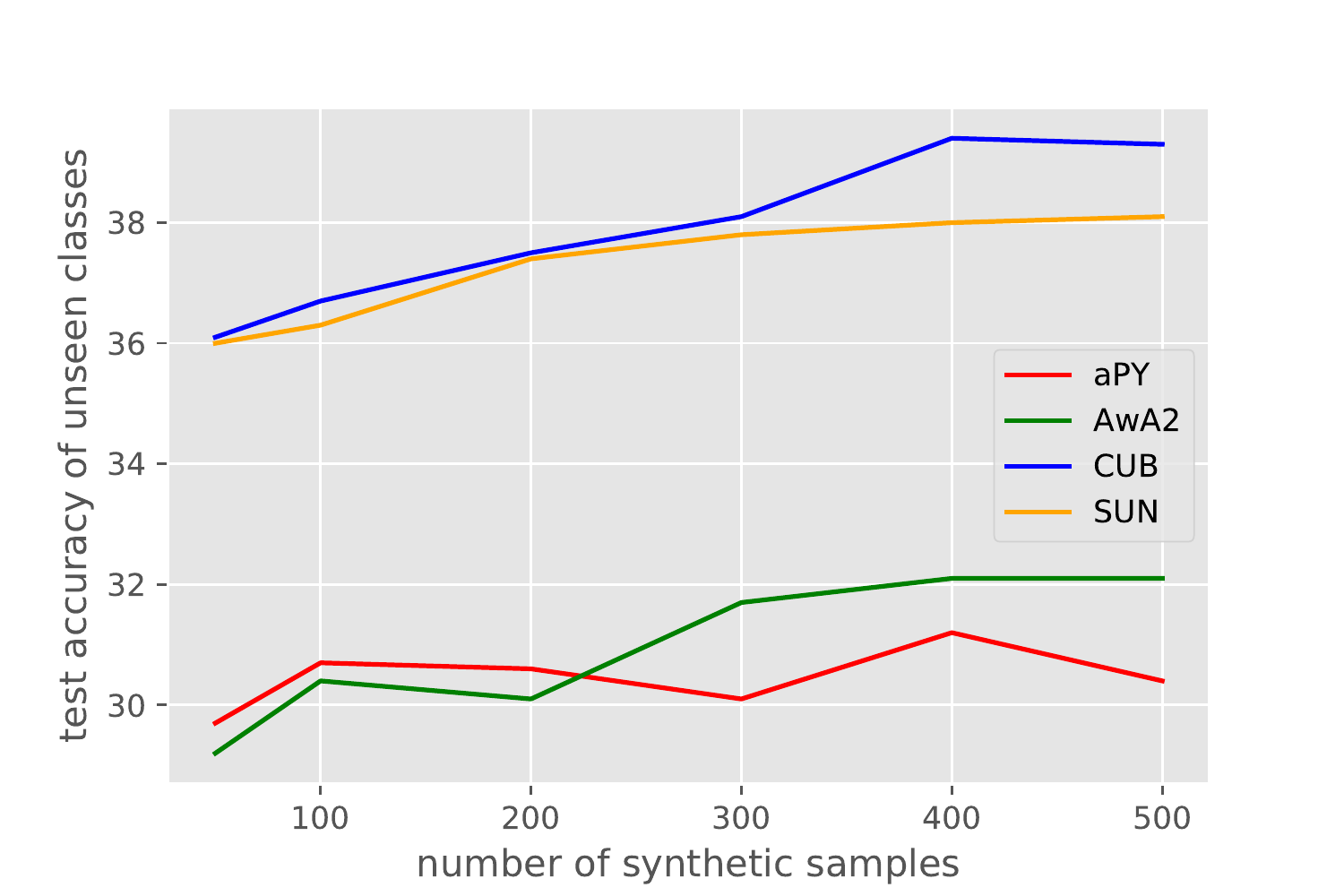}
\end{center}
   \caption{Test accuracy for unseen classes with respect to  the number of synthetic samples.}
\label{fig:test-samples}
\end{figure}

\begin{figure*}[t]
\centering
\subfigure[]{
\begin{minipage}[l]{0.4\linewidth}
\centering
\includegraphics[width=1\textwidth]{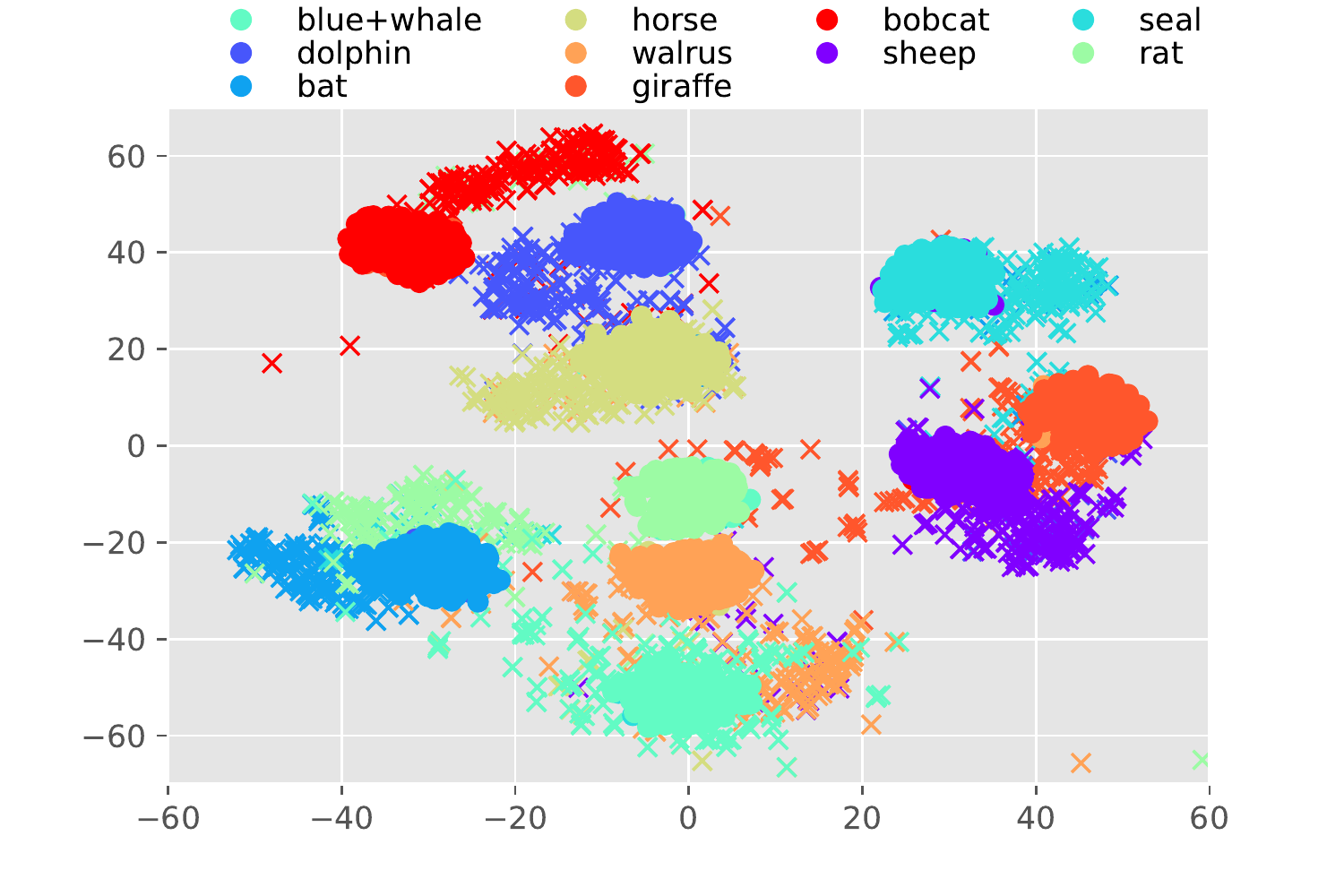}
\end{minipage}
\label{fig:vis_awa}
}
\subfigure[]{
\begin{minipage}[l]{0.4\linewidth}
\centering
\includegraphics[width=1\textwidth]{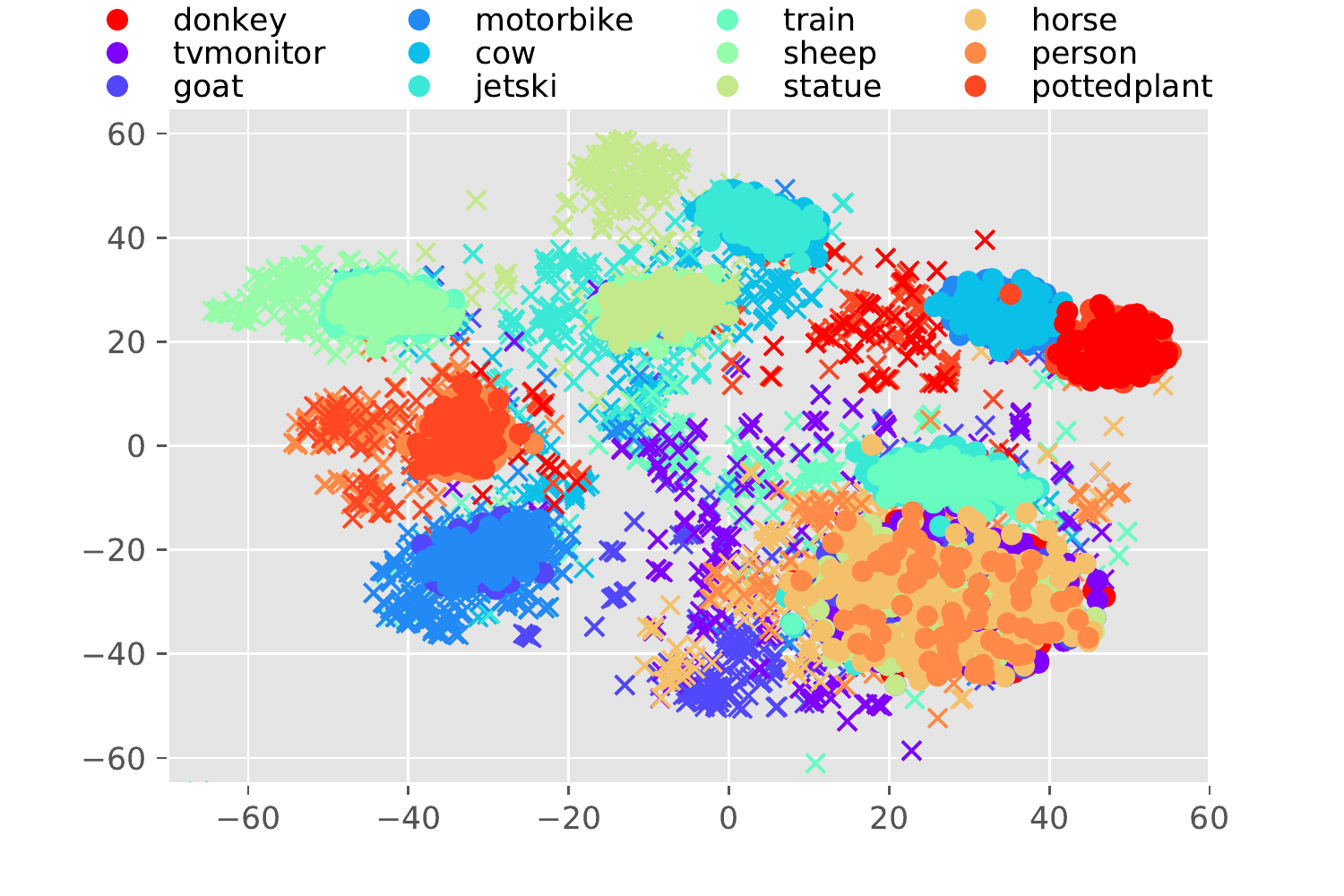}
\end{minipage}
\label{fig:vis_apy}
}
\caption{t-SNE visualization of synthetic ($\bullet$) and real ($\times$) image features  for unseen classes in (a) AwA2 and (b) aPY datasets.}
\label{fig:vis}
\end{figure*}

\subsection{Effect of the Number of Synthetic Samples}
We also analyze how the number of synthetic samples affects test accuracy, as shown in Figure~\ref{fig:test-samples}. As we can see, for AwA2~\cite{xian2018gbu}, increasing the number of synthesized examples does not help much in increasing the prediction accuracy on unseen classes, while for the other classes, the accuracy increases as the number of synthesized samples increases, and saturates when the number of synthetic samples is about 400. Also, CUB~\cite{wah2011cub_bird} and SUN~\cite{patterson2012sun} have a larger improvement in performance as the number of synthetic samples increases than the other two datasets, which may be because of there are much more unseen classes in CUB~\cite{wah2011cub_bird} (50) and SUN~\cite{patterson2012sun} (72) than that of aPY~\cite{farhadi2009apy} (12) and AwA2~\cite{xian2018gbu} (10), and thus more synthetic samples can better help distinguishing images among those classes.

\subsection{Visualization of Synthetic Image Features}

In order to provide a qualitative evaluation on our proposed \model~ model, we visualize some synthetic image features along with some real image features, and the results are illustrated in Figure~\ref{fig:vis}. Since the numbers of unseen classes for CUB~\cite{wah2011cub_bird} and SUN~\cite{patterson2012sun} are relatively large and thus hard to visualize, we only visualize the results of AwA2~\cite{xian2018gbu} and aPY~\cite{farhadi2009apy}. For each unseen class we synthesize 200 image features, and we also randomly sample 200 real image features for each unseen class as well, and then we use t-SNE~\cite{maaten2008tsne} to reduce the dimension to two for visualization.

From the real samples in Figure~\ref{fig:vis_awa} we can see that some classes overlap with each other by a large degree, such as \emph{blue whale} and \emph{walrus}, \emph{rat} and \emph{bat}. This overlapping is reasonable since \emph{bat} and \emph{rat} are biologically similar to each other, and thus it make sense for the synthetic samples of \emph{bat} to be close to \emph{rat} as well.  For most classes like \emph{bobcat}, \emph{horse}, \emph{dolphin}, the synthetic image features are very close to the true samples, where some even overlap with true samples quite well, such as \emph{blue whale}, \emph{sheep} and \emph{giraffe}. One failure case is the \emph{rat} class, where we can see that the synthetic samples are far from the true samples. Except for this, as we can see from Figure~\ref{fig:vis_awa}, 1-NN can predict the label of test images quite well for most classes.

Similar conclusions can be drawn for aPY~\cite{farhadi2009apy} as well, as  shown in Figure~\ref{fig:vis_apy}. It should be noted that the real samples of aPY~\cite{farhadi2009apy} are  densely  cluttered and that many classes overlap with each other by a large degree, especially in the down-right part of Figure~\ref{fig:vis_apy}, where there are at least 4 classes almost completely overlap together. From the visualization of other classes that are put closer to the boundary of the figure, we can see that our model generates very good visual features that matches the real ones quite well, such as the \emph{sheep}, \emph{motorbike} and \emph{person} classes. Even for the densely cluttered area, the synthetic features still totally lie within that clutter, which means our model can still generate very good examples of those cluttered classes. 
Just as the AwA2~\cite{xian2018gbu} dataset, however, there are still some failure cases here, for example, the synthetic features of the \emph{statue} class are not very close to the real features.



\section{Conclusion}
In this paper, we study the generalized zero-shot learning problem and propose \model, a model that unifies three different approaches:  $visual\rightarrow semantic$ mapping (generator), $semantic\rightarrow visual$ mapping (regressor), and metric learning (discriminator). 
The generator and the regressor learn from each other in a dual learning fashion, while they both learn from the discriminator through a dual adversarial loss. In this way, our model provides a unified framework to bridge the gap between visual and semantic space in a generative dual adversarial framework. 
Extensive experiments on four benchmark datasets demonstrate the effectiveness of our model in balancing accuracy between seen and unseen classes. Component analysis also show that each of the three components can benefit from jointly training together and thus demonstrate the effectiveness of our proposed model. We also visualize the synthetic visual features of unseen classes to show that our model is able to generate high quality visual features.


{\small
\bibliographystyle{ieee}
\bibliography{reference}

\begin{thebibliography}{10}\itemsep=-1pt

\bibitem{akata2013ALE}
Z.~Akata, F.~Perronnin, Z.~Harchaoui, and C.~Schmid.
\newblock Label-embedding for attribute-based classification.
\newblock In {\em Proceedings of the IEEE Conference on Computer Vision and
  Pattern Recognition}, pages 819--826, 2013.

\bibitem{akata2015SJE}
Z.~Akata, S.~Reed, D.~Walter, H.~Lee, and B.~Schiele.
\newblock Evaluation of output embeddings for fine-grained image
  classification.
\newblock In {\em Proceedings of the IEEE Conference on Computer Vision and
  Pattern Recognition}, pages 2927--2936, 2015.

\bibitem{annadani2018psr}
Y.~Annadani and S.~Biswas.
\newblock Preserving semantic relations for zero-shot learning.
\newblock In {\em Proceedings of the IEEE Conference on Computer Vision and
  Pattern Recognition}, pages 7603--7612, 2018.

\bibitem{arjovsky2017wgan}
M.~Arjovsky, S.~Chintala, and L.~Bottou.
\newblock Wasserstein gan.
\newblock {\em arXiv preprint arXiv:1701.07875}, 2017.

\bibitem{changpinyo2016SYNC}
S.~Changpinyo, W.-L. Chao, B.~Gong, and F.~Sha.
\newblock Synthesized classifiers for zero-shot learning.
\newblock In {\em Proceedings of the IEEE Conference on Computer Vision and
  Pattern Recognition}, pages 5327--5336, 2016.

\bibitem{chao2016empirical}
W.-L. Chao, S.~Changpinyo, B.~Gong, and F.~Sha.
\newblock An empirical study and analysis of generalized zero-shot learning for
  object recognition in the wild.
\newblock In {\em European Conference on Computer Vision}, pages 52--68.
  Springer, 2016.

\bibitem{chen2018SPAEN}
L.~Chen, H.~Zhang, J.~Xiao, W.~Liu, and S.-F. Chang.
\newblock Zero-shot visual recognition using semantics-preserving adversarial
  embedding network.
\newblock In {\em Proceedings of the IEEE Conference on Computer Vision and
  Pattern Recognition}, volume~2, 2018.

\bibitem{chen2016variational}
X.~Chen, D.~P. Kingma, T.~Salimans, Y.~Duan, P.~Dhariwal, J.~Schulman,
  I.~Sutskever, and P.~Abbeel.
\newblock Variational lossy autoencoder.
\newblock {\em arXiv preprint arXiv:1611.02731}, 2016.

\bibitem{dinu2014improving}
G.~Dinu, A.~Lazaridou, and M.~Baroni.
\newblock Improving zero-shot learning by mitigating the hubness problem.
\newblock {\em arXiv preprint arXiv:1412.6568}, 2014.

\bibitem{farhadi2009apy}
A.~Farhadi, I.~Endres, D.~Hoiem, and D.~Forsyth.
\newblock Describing objects by their attributes.
\newblock In {\em Computer Vision and Pattern Recognition, 2009. CVPR 2009.
  IEEE Conference on}, pages 1778--1785. IEEE, 2009.

\bibitem{frome2013devise}
A.~Frome, G.~S. Corrado, J.~Shlens, S.~Bengio, J.~Dean, T.~Mikolov, et~al.
\newblock Devise: A deep visual-semantic embedding model.
\newblock In {\em Advances in neural information processing systems}, pages
  2121--2129, 2013.

\bibitem{goodfellow2014GAN}
I.~Goodfellow, J.~Pouget-Abadie, M.~Mirza, B.~Xu, D.~Warde-Farley, S.~Ozair,
  A.~Courville, and Y.~Bengio.
\newblock Generative adversarial nets.
\newblock In {\em Advances in neural information processing systems}, pages
  2672--2680, 2014.

\bibitem{gulrajani2017wgan-gp}
I.~Gulrajani, F.~Ahmed, M.~Arjovsky, V.~Dumoulin, and A.~C. Courville.
\newblock Improved training of wasserstein gans.
\newblock In {\em Advances in Neural Information Processing Systems}, pages
  5767--5777, 2017.

\bibitem{he2016duallearning}
D.~He, Y.~Xia, T.~Qin, L.~Wang, N.~Yu, T.~Liu, and W.-Y. Ma.
\newblock Dual learning for machine translation.
\newblock In {\em Advances in Neural Information Processing Systems}, pages
  820--828, 2016.

\bibitem{kingma2014adam}
D.~P. Kingma and J.~Ba.
\newblock Adam: A method for stochastic optimization.
\newblock {\em arXiv preprint arXiv:1412.6980}, 2014.

\bibitem{kingma2013vae}
D.~P. Kingma and M.~Welling.
\newblock Auto-encoding variational bayes.
\newblock {\em arXiv preprint arXiv:1312.6114}, 2013.

\bibitem{kodirov2017SAE}
E.~Kodirov, T.~Xiang, and S.~Gong.
\newblock Semantic autoencoder for zero-shot learning.
\newblock {\em arXiv preprint arXiv:1704.08345}, 2017.

\bibitem{lampert2009DAP-IAP}
C.~H. Lampert, H.~Nickisch, and S.~Harmeling.
\newblock Learning to detect unseen object classes by between-class attribute
  transfer.
\newblock In {\em Computer Vision and Pattern Recognition, 2009. CVPR 2009.
  IEEE Conference on}, pages 951--958. IEEE, 2009.

\bibitem{luo2017deep}
P.~Luo, G.~Wang, L.~Lin, and X.~Wang.
\newblock Deep dual learning for semantic image segmentation.
\newblock In {\em Proceedings of the IEEE Conference on Computer Vision and
  Pattern Recognition, Honolulu, HI, USA}, pages 21--26, 2017.

\bibitem{maaten2008tsne}
L.~v.~d. Maaten and G.~Hinton.
\newblock Visualizing data using t-sne.
\newblock {\em Journal of machine learning research}, 9(Nov):2579--2605, 2008.

\bibitem{makhzani2015aae}
A.~Makhzani, J.~Shlens, N.~Jaitly, I.~Goodfellow, and B.~Frey.
\newblock Adversarial autoencoders.
\newblock {\em arXiv preprint arXiv:1511.05644}, 2015.

\bibitem{mao2017lsgan}
X.~Mao, Q.~Li, H.~Xie, R.~Y. Lau, Z.~Wang, and S.~P. Smolley.
\newblock Least squares generative adversarial networks.
\newblock In {\em Computer Vision (ICCV), 2017 IEEE International Conference
  on}, pages 2813--2821. IEEE, 2017.

\bibitem{mishra2017cvae-zsl}
A.~Mishra, M.~Reddy, A.~Mittal, and H.~A. Murthy.
\newblock A generative model for zero shot learning using conditional
  variational autoencoders.
\newblock {\em arXiv preprint arXiv:1709.00663}, 2017.

\bibitem{patterson2012sun}
G.~Patterson and J.~Hays.
\newblock Sun attribute database: Discovering, annotating, and recognizing
  scene attributes.
\newblock In {\em Computer Vision and Pattern Recognition (CVPR), 2012 IEEE
  Conference on}, pages 2751--2758. IEEE, 2012.

\bibitem{radford2015dcgan}
A.~Radford, L.~Metz, and S.~Chintala.
\newblock Unsupervised representation learning with deep convolutional
  generative adversarial networks.
\newblock {\em arXiv preprint arXiv:1511.06434}, 2015.

\bibitem{reed2016txt2img}
S.~Reed, Z.~Akata, X.~Yan, L.~Logeswaran, B.~Schiele, and H.~Lee.
\newblock Generative adversarial text to image synthesis.
\newblock {\em arXiv preprint arXiv:1605.05396}, 2016.

\bibitem{romera2015eszsl}
B.~Romera-Paredes and P.~Torr.
\newblock An embarrassingly simple approach to zero-shot learning.
\newblock In {\em International Conference on Machine Learning}, pages
  2152--2161, 2015.

\bibitem{ILSVRC15ImageNet}
O.~Russakovsky, J.~Deng, H.~Su, J.~Krause, S.~Satheesh, S.~Ma, Z.~Huang,
  A.~Karpathy, A.~Khosla, M.~Bernstein, A.~C. Berg, and L.~Fei-Fei.
\newblock {ImageNet Large Scale Visual Recognition Challenge}.
\newblock {\em International Journal of Computer Vision (IJCV)},
  115(3):211--252, 2015.

\bibitem{shigeto2015ridge}
Y.~Shigeto, I.~Suzuki, K.~Hara, M.~Shimbo, and Y.~Matsumoto.
\newblock Ridge regression, hubness, and zero-shot learning.
\newblock In {\em Joint European Conference on Machine Learning and Knowledge
  Discovery in Databases}, pages 135--151. Springer, 2015.

\bibitem{sohn2015cvae}
K.~Sohn, H.~Lee, and X.~Yan.
\newblock Learning structured output representation using deep conditional
  generative models.
\newblock In {\em Advances in Neural Information Processing Systems}, pages
  3483--3491, 2015.

\bibitem{verma2018segzsl}
V.~K. Verma, G.~Arora, A.~Mishra, and P.~Rai.
\newblock Generalized zero-shot learning via synthesized examples.
\newblock In {\em Proceedings of the IEEE Conference on Computer Vision and
  Pattern Recognition (CVPR)}, 2018.

\bibitem{wah2011cub_bird}
C.~Wah, S.~Branson, P.~Welinder, P.~Perona, and S.~Belongie.
\newblock The caltech-ucsd birds-200-2011 dataset.
\newblock 2011.

\bibitem{xian2016LATEM}
Y.~Xian, Z.~Akata, G.~Sharma, Q.~Nguyen, M.~Hein, and B.~Schiele.
\newblock Latent embeddings for zero-shot classification.
\newblock In {\em Proceedings of the IEEE Conference on Computer Vision and
  Pattern Recognition}, pages 69--77, 2016.

\bibitem{xian2018gbu}
Y.~Xian, C.~H. Lampert, B.~Schiele, and Z.~Akata.
\newblock Zero-shot learning-a comprehensive evaluation of the good, the bad
  and the ugly.
\newblock {\em IEEE transactions on pattern analysis and machine intelligence},
  2018.

\bibitem{xian2018fganzsl}
Y.~Xian, T.~Lorenz, B.~Schiele, and Z.~Akata.
\newblock Feature generating networks for zero-shot learning.
\newblock In {\em Proceedings of the IEEE conference on computer vision and
  pattern recognition}, 2018.

\bibitem{yang2018relation}
F.~S.~Y. Yang, L.~Zhang, T.~Xiang, P.~H. Torr, and T.~M. Hospedales.
\newblock Learning to compare: Relation network for few-shot learning.
\newblock 2018.

\bibitem{yi2017dualgan}
Z.~Yi, H.~R. Zhang, P.~Tan, and M.~Gong.
\newblock Dualgan: Unsupervised dual learning for image-to-image translation.
\newblock In {\em ICCV}, pages 2868--2876, 2017.

\bibitem{zhang2017DEM}
L.~Zhang, T.~Xiang, S.~Gong, et~al.
\newblock Learning a deep embedding model for zero-shot learning.
\newblock 2017.

\bibitem{zhang2015SSE}
Z.~Zhang and V.~Saligrama.
\newblock Zero-shot learning via semantic similarity embedding.
\newblock In {\em Proceedings of the IEEE international conference on computer
  vision}, pages 4166--4174, 2015.

\bibitem{zhao2017infovae}
S.~Zhao, J.~Song, and S.~Ermon.
\newblock Infovae: Information maximizing variational autoencoders.
\newblock {\em arXiv preprint arXiv:1706.02262}, 2017.

\bibitem{zhao2018unsupervised}
T.~Zhao, K.~Lee, and M.~Eskenazi.
\newblock Unsupervised discrete sentence representation learning for
  interpretable neural dialog generation.
\newblock {\em arXiv preprint arXiv:1804.08069}, 2018.

\bibitem{zhu2017cyclegan}
J.-Y. Zhu, T.~Park, P.~Isola, and A.~A. Efros.
\newblock Unpaired image-to-image translation using cycle-consistent
  adversarial networks.
\newblock {\em arXiv preprint}, 2017.

\end{thebibliography}
}

\end{document}